# LLM-Match: An Open-Sourced Patient Matching Model Based on Large Language Models and Retrieval-Augmented Generation

Xiaodi Li, Ph.D.[1], Shaika Chowdhury, Ph.D.[1], Chung Il Wi, M.D.[2], Maria Vassilaki, M.D., Ph.D.[3], Xiaoke Liu, M.D., Ph.D.[4], Terence T Sio, M.D., M.S.[5], Owen Garrick, M.D.[6], Young J Juhn, M.D., M.P.H.[2], James R Cerhan, M.D. Ph.D.[7], Cui Tao, Ph.D.[1], Nansu Zong, Ph.D.[1, *]

[1]Department of Artificial Intelligence and Informatics Research, Mayo Clinic, Rochester, MN, USA; [2]Department of Pediatric and Adolescent Medicine, Mayo Clinic, Rochester, MN, USA; [3]Department of Quantitative Health Sciences, Mayo Clinic, Rochester, MN, USA; [4]Cardiology Department, Mayo Clinic Health System, La Crosse, WI, USA; [5]Department of Radiation Oncology, Mayo Clinic, Rochester, MN, USA; [6]Clinical Trials, Mayo Clinic, Rochester, MN, USA; [7]Department of Quantitative Health Sciences, Mayo Clinic, Rochester, MN, USA

[*] Corresponding author, Nansu Zong (zong.nansu@mayo.edu).

**Abstract**

*Patient matching is the process of linking patients to appropriate clinical trials by accurately identifying and matching their medical records with trial eligibility criteria. We propose LLM-Match, a novel framework for patient matching leveraging fine-tuned open-source large language models. Our approach consists of four key components. First, a retrieval-augmented generation (RAG) module extracts relevant patient context from a vast pool of electronic health records (EHRs). Second, a prompt generation module constructs input prompts by integrating trial eligibility criteria (both inclusion and exclusion criteria), patient context, and system instructions. Third, a fine-tuning module with a classification head optimizes the model parameters using structured prompts and ground-truth labels. Fourth, an evaluation module assesses the fine-tuned model's performance on the testing datasets. We evaluated LLM-Match on four open datasets—n2c2, SIGIR, TREC 2021, and TREC 2022—using open-source models, comparing it against TrialGPT, Zero-Shot, and GPT-4-based closed models. LLM-Match outperformed all baselines.*

**Introduction**

Matching patients to suitable clinical trials is a critical yet challenging task in clinical research. Clinical trials are essential for evaluating new treatments and advancing medical science, but patient recruitment remains a significant bottleneck, often leading to trial delays or failures[1]. Traditional recruitment methods rely on manual screening, where clinical coordinators review electronic health records (EHRs) against complex eligibility criteria—a time-consuming and error-prone process[2]. Studies have shown that up to 80% of clinical trials fail to meet enrollment timelines due to inefficient matching methods, causing significant financial and scientific setbacks[3]. The vast volume of unstructured clinical text within EHRs, including physician notes, diagnostic reports, and medication histories, further complicates this task[4]. Automating patient-trial matching is therefore a crucial step toward improving efficiency and ensuring broader access to clinical studies.

Early approaches to clinical trial patient matching focused on rule-based systems that converted unstructured eligibility criteria into structured queries executable over electronic health records (EHRs)[5]. Tools like EliXR applied deterministic rules to extract criteria, while later methods such as EliIE and Criteria2Query integrated natural language processing (NLP) techniques, including named entity recognition (NER) and relation extraction, to improve accuracy[6, 7]. The RBC model demonstrated strong performance in structured cohort selection tasks but remained limited in its ability to generalize across diverse trial requirements[8].

Despite their effectiveness in specific cases, these systems often required extensive maintenance and struggled with adaptability[9]. This led to the exploration of deep learning models for patient-trial matching. Neural network-based solutions such as COMPOSE and DeepEnroll encoded eligibility criteria alongside structured EHR data, offering improved generalization[10, 11]. However, these methods primarily relied on structured information, neglecting the wealth of insights available in unstructured clinical text.

With the rise of large language models (LLMs), new approaches sought to leverage their capabilities in processing free-text eligibility criteria. TrialGPT pioneered the use of LLMs for direct patient-trial matching, achieving high accuracy but lacking a zero-shot evaluation component[12]. Scaling with LLMs expanded on this by using GPT-4[13] to convert free-text criteria into structured queries[14]. TRIALSCOPE introduced a broader concept of real-world evidence for trial simulation but did not include a dedicated patient-matching framework[15]. TrialLlama fine-tuned an open-source LLM for trial matching but required extensive computational resources, limiting its practicality[16, 17].

Recent advances in large language models (LLMs) have shown promise in automating patient-trial matching by leveraging natural language processing (NLP) to analyze unstructured clinical text[18]. Unlike rule-based or structured query approaches, LLMs can process free-text eligibility criteria and patient information, improving flexibility and generalization[19]. However, a key challenge lies in efficiently retrieving relevant patient context from extensive EHRs while maintaining model accuracy and interpretability. Directly processing entire patient records is computationally expensive and impractical due to context length limitations[20]. Prior work has attempted to address this challenge using specialized clinical NLP models, such as BioBERT and MedBERT, but these models often struggle with long-text dependencies and domain-specific adaptation[21, 22].

Additionally, retrieval-augmented generation (RAG) has been explored as a solution for grounding model outputs with external knowledge sources, mitigating hallucination issues and improving contextual understanding in medical applications[12]. Existing works on retrieval-augmented generation (RAG) are largely based on commercial LLMs like GPT-4[13], which come with privacy and data security concerns. Using such models requires sharing sensitive medical data externally, posing risks to patient health information protection and limiting their applicability in privacy-critical settings. Reliance on commercial models means limited control over data storage, access, and model updates, potentially introducing biases or ethical concerns when applied to sensitive healthcare applications. These privacy risks make it essential to explore secure, open-source, or on-premise RAG implementations that ensure data sovereignty while leveraging external knowledge effectively. Moreover, current works are based on the generative head, limited by its tendency to produce open-ended, inconsistent, or ambiguous outputs, making it less suitable for tasks requiring structured decision-making. Its reliance on free-form text generation can lead to unpredictable responses, reducing precision and interpretability, especially in applications that demand clear decision boundaries. Furthermore, generative models often require significant computational resources and extensive fine-tuning to align outputs with specific tasks, further complicating their deployment in structured classification problems.

To address these challenges, we propose LLM-Match, a novel framework that fine-tunes open-source LLMs for patient-trial matching. Our method consists of four key components: (1) a retrieval-augmented generation (RAG) module that extracts the most relevant patient information from large-scale EHRs, (2) a prompt generation module that constructs structured inputs integrating trial eligibility criteria, patient data, and system instructions, (3) a fine-tuning module with a classification head that optimizes model performance using structured prompts and ground-truth labels, and (4) an evaluation module that assesses model effectiveness on benchmark datasets. We evaluate LLM-Match on four publicly available datasets, the n2c2 2018 cohort selection benchmark[23], the Special Interest Group on Information Retrieval (SIGIR) in 2016[24], and the 2021 and 2022 Clinical Trials (CT) tracks[25, 26] of the Text REtrieval Conference (TREC), achieving strong results in patient-trial eligibility prediction. Our findings demonstrate that integrating retrieval-augmented generation with fine-tuned LLMs significantly improves patient-matching accuracy compared to standard LLM-based methods.

Our contributions are fourfold: (1) we develop LLM-Match, a fine-tuned LLM framework tailored for patient-trial matching, enhanced with a classification head for precise eligibility predictions, (2) we integrate retrieval-augmented generation (RAG) to efficiently process unstructured EHR data, (3) we utilize small open-sources models, such as LLaMA 3[27] (8B), offering a transparent and adaptable alternative to methods relying on closed-source models like GPT-4[13], and (4) we conduct comprehensive evaluations that establish state-of-the-art performance on benchmark datasets. These contributions highlight the potential of LLM-Match to transform clinical trial recruitment by providing a scalable, automated, and accurate solution for patient-trial matching. The novelty of the framework lies in the fine-tuning module, which includes a classification head. Additionally, our method leverages small open-source models, such as LLaMA 3[27] (8B), in contrast to other approaches that rely on closed-source model like GPT-4[13]. By incorporating retrieval-based selection alongside LLM-based reasoning, our approach enhances both accuracy and efficiency in patient-trial matching.

## Methods

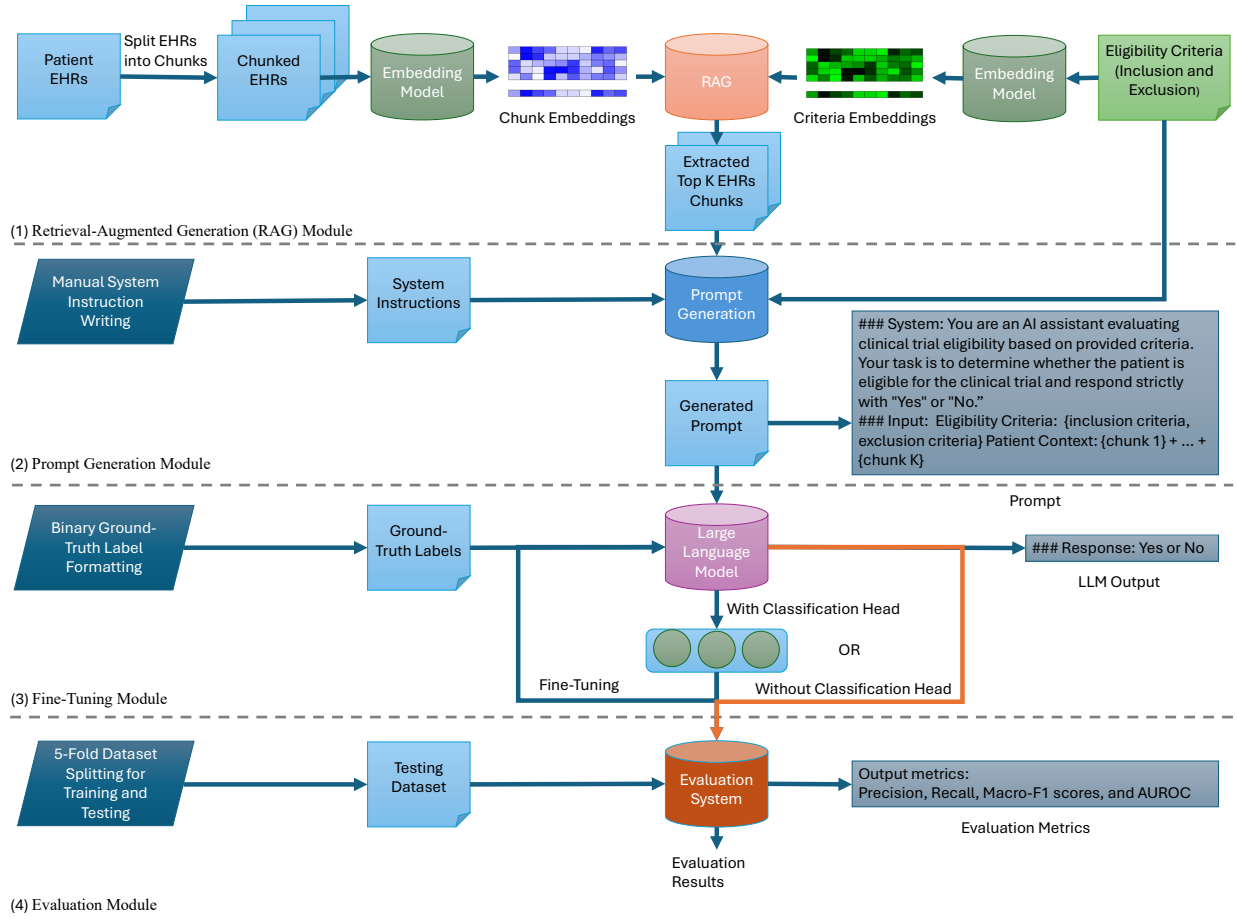

**Figure 1.** Overview of LLM-Match. First, we divide the patient EHRs into smaller chunks. Then, we independently encode the chunked EHRs and eligibility criteria using an embedding model, specifically BioBERT[28] in our case. The RAG module retrieves the top-k most relevant electronic health records (EHRs) based on inclusion and exclusion criteria (k = 4) by measuring the similarity between chunk embeddings and criteria embeddings. The prompt generation module then integrates the retrieved EHRs and eligibility criteria with predefined system instructions to construct a structured input prompt. Next, the fine-tuning module trains a large language model (LLaMA 3[27] using the generated prompts and ground-truth labels, employing two approaches: (1) the original model and (2) the original model with an added classification head. Finally, the evaluation module measures the fine-tuned model's performance on a testing dataset to assess its effectiveness in patient-trial matching.

Figure 1 provides an overview of our proposed LLM-Match framework, designed for efficient and accurate patient-trial matching. The framework consists of four key modules that work together to process unstructured EHR data and eligibility criteria, ensuring robust performance in real-world clinical settings: _Retrieval-Augmented Generation (RAG) Module_: This module retrieves the most relevant patient records from large-scale unstructured EHR data, ensuring that only pertinent information is fed into the LLM. It enhances model efficiency by filtering irrelevant data and providing contextualized input. _Prompt Generation Module_: To facilitate effective patient-trial matching, this module dynamically constructs structured prompts based on the retrieved clinical data and eligibility criteria. By leveraging domain-specific templates and contextual embeddings, it ensures optimal LLM comprehension and response generation. _Fine-Tuning Module_: This module adapts the LLM to patient-trial matching tasks using domain-specific datasets. It includes supervised fine-tuning with labeled clinical data and incorporates a classification head for eligibility prediction, enhancing interpretability and decision-making. _Evaluation Module_: To validate model performance, this module employs benchmark datasets and real-world clinical trial data. It assesses effectiveness using key metrics such as precision, recall, Macro-F1, and AUROC, ensuring reliable and scalable patient-trial matching.

As shown in Figure 1, we split the patient EHRs into multiple chunks and encode them using an embedding model, specifically BioBERT[28] in our case, to obtain chunk embeddings. Similarly, we encode the eligibility criteria using the same embedding model to generate the criteria embedding. Then, Retrieval-Augmented Generation (RAG) module retrieves the top-k most relevant electronic health records (EHRs) based on the given eligibility criteria, which include both inclusion and exclusion conditions, by measuring the similarity between chunk embeddings and criteria embeddings. This retrieval process leverages similarity-based search techniques to ensure that the most pertinent clinical notes are selected, effectively reducing noise and improving matching accuracy. In our implementation, we typically set k = 4, balancing efficiency and informativeness.

Next, the Prompt Generation Module integrates the retrieved EHRs and eligibility criteria with predefined system instructions to construct a structured input prompt suitable for processing by a large language model (LLM). This module ensures that the information is presented in a coherent and contextually rich format, facilitating more precise and relevant LLM responses. It incorporates domain-specific templates, optimizes prompt formatting, and applies instruction tuning to enhance model comprehension and reasoning over complex clinical narratives.

Then, the Fine-Tuning Module adapts a large language model (LLM)—in our case, LLaMA 3[27]—by training it on generated prompts paired with ground-truth labels as supervision. This ensures that the model learns to correctly match patients to trials based on real-world examples. We implement fine-tuning using two complementary approaches: (1) standard fine-tuning, which optimizes the original model parameters for the task, and (2) fine-tuning with a classification head, which extends the base model by adding a task-specific layer to explicitly predict patient-trial eligibility, improving interpretability and structured decision-making.

Finally, the Evaluation Module rigorously assesses the fine-tuned model's performance using a dedicated testing dataset, ensuring its effectiveness in patient-trial matching. This evaluation process includes benchmarking against established performance metrics such as precision, recall, Macro-F1, and AUROC along with ablation studies to analyze the impact of each module. Additionally, we conduct robustness testing across different patient cohorts to assess generalizability, ensuring that LLM-Match remains adaptable across diverse clinical trial settings.

By integrating retrieval, structured prompting, fine-tuning, and evaluation, our LLM-Match framework offers a scalable, efficient, and high-accuracy solution for automating patient-trial matching, significantly reducing the manual effort required in clinical trial recruitment.

The following are the detailed formulas corresponding to different modules.

*Retrieval-Augmented Generation (RAG) Module*
Given a set of patients $P$ and their corresponding electronic health records (EHRs) $N$, where each patient $p \in P$ has a set of notes $N_p \subseteq N$, our objective is to identify the top-k most relevant records based on the given trial eligibility criteria $C$. The eligibility criteria contain both inclusion and exclusion conditions, denoted as $C = \{c_1, c_2, \ldots, c_m\}$. To efficiently retrieve relevant patient context, we embed each EHR note and trial criterion using an embedding model $M$, which maps textual data to a vector space, shown as $M: T \to R^d$, where $T$ represents the text corpus, and $d$ is the dimensionality of the embedding space. For each criterion $c$, we compute its embedding $v_c$, and for each note $n$ of a patient $p$, we compute its embedding $v_n$ as $v_c = M(c), v_n = M(n)$. To rank the relevance of patient notes to the trial criteria, we compute the cosine similarity between each note embedding and the criterion embedding, denoted as $Sim(v_n, v_c) = \frac{v_n \cdot v_c}{\|v_n\|\|v_c\|}$. The top-k most relevant notes are selected based on their similarity scores, shown as $N_p^{(k)} = argmax_{N_p} \sum_{c \in C} Sim(v_n, v_c)$, where $N_p^{(k)}$ represents the top-k selected EHR notes for patient $p$. In our experiments, we set $k = 4$ based on empirical validation.

*Prompt Generation Module*
Once the top-k retrieved EHR notes are selected, they are combined with trial eligibility criteria and system instructions to construct a structured input prompt. The prompt is defined as $\Phi(p) = System\ Instructions \oplus C \oplus N_p^{(k)}$, where $\oplus$ denotes concatenation. The system instructions provide guidance on how the large language model

(LLM) should interpret the patient records and eligibility criteria. This structured prompt ensures that the fine-tuning model receives only the most relevant patient context, reducing unnecessary computational overhead.

*Fine-Tuning Module*
We fine-tune an open-source large language model (LLaMA 3[27]) to learn patient-trial matching based on the structured prompts generated in the previous step. Let $\theta$ represent the parameters of the LLM and let $D$ denote the training dataset. Each training sample consists of an input prompt $\Phi(p)$ and a ground-truth label $y_p$ indicating whether the patient meets the eligibility criteria, denoted as $D = \{(\Phi(p), y_p)\}_{p \in P}$. The fine-tuning process optimizes the model parameters $\theta$ by minimizing a binary cross-entropy loss function[29], shown as $L(\theta) = -\sum_{p \in P} [y_p \log f_\theta(\Phi(p)) + (1 - y_p) \log(1 - f_\theta(\Phi(p)))]$, where $f_\theta(\Phi(p))$ is the model's predicted probability that patient $p$ meets the eligibility criteria. We employ gradient-based optimization **(e.g., Adam optimizer)** for fine-tuning, denoted as $\theta^{t+1} = \theta^t - \eta \nabla_\theta L(\theta^t)$, where $\eta$ is the learning rate. The fine-tuning process continues until convergence, ensuring the model generalizes well to unseen patients.

*Evaluation Module*
We evaluate our models based on multiple performance metrics to comprehensively assess their effectiveness in patient-trial matching. Our primary evaluation focuses on precision, recall, and Macro-F1 scores for the binary classification task of determining whether a patient meets each of the 13 inclusion criteria in the 2018 n2c2 benchmark[23]. These metrics provide insights into the model's ability to accurately classify patients while balancing both sensitivity and specificity.

Following the official n2c2 evaluation protocol, we compute the overall performance score as the simple average of the F1 scores for both the "met" and "not met" classifications. This ensures consistency with prior work, enabling direct comparisons with existing patient-trial matching models. By averaging across both classes, we mitigate potential biases introduced by class imbalance, ensuring a fair evaluation.

Beyond the n2c2 benchmark, we extend our evaluation to SIGIR[24], TREC 2021[25], and TREC 2022[26] datasets, where we report additional metrics to capture model robustness across different clinical datasets. Specifically, we compute the Area Under the Receiver Operating Characteristic Curve (AUROC)[30] to assess the model's ability to distinguish between positive and negative cases across varying decision thresholds. AUROC serves as a strong indicator of a model's discriminative power, particularly in real-world clinical settings where decision boundaries may shift.

Precision[31] measures the proportion of correctly identified positive cases among all predicted positive cases, Recall[31] measures the proportion of correctly identified positive cases among all actual positive cases, F1-score[31] is the harmonic mean of precision and recall. Macro-F1[32] is computed as the simple average of the F1-score for the "met" class and the F1-score for the "not met" class, treating them equally. This ensures balanced evaluation across both positive and negative classifications. Area Under the Receiver Operating Characteristic Curve (AUROC)[30] measures the ability of the model to distinguish between positive and negative cases across different decision thresholds. It is computed as the area under the ROC curve, which plots the true positive rate (recall) against the false positive rate. A higher AUROC indicates better discriminative performance.

**Table 1.** Datasets Statistics.

| Dataset | # of Patients | # of Trials | Patient-Trial Pairs |
| --- | --- | --- | --- |
| n2c2 | 202 | 13 criteria | 2626 |
| SIGIR | 58 | 2991 | 3141 |
| TREC 2021 | 75 | 13142 | 15437 |
| TREC 2022 | 50 | 11604 | 12682 |

*Datasets*
<u>n2c2</u>: The 2018 n2c2 Clinical Trial Cohort Selection dataset[23] is the largest publicly available benchmark for clinical trial patient matching. It consists of 288 diabetic patients (202 train, 86 test), each with 2-5 de-identified clinical notes averaging 2,711 words. Patients were sourced from Mass General and Brigham and Women's hospitals. The dataset includes 13 inclusion criteria, commonly used in eligibility assessments, but no exclusion criteria. Matching is performed using natural language processing as all information is unstructured text. Labels were annotated by two medical experts, determining whether each criterion was "MET" or "NOT MET", with an inter-annotator agreement (Cohen's Kappa) of 0.54. This dataset simulates a synthetic trial rather than real clinical enrollments.

**Table 2.** Experimental Results on the n2c2 2018 Cohort Selection Challenge.

| Model | Precision | Recall | Overall Macro-F1 |
| --- | --- | --- | --- |
| Basic Baseline | 0.69 | 0.78 | 0.43 |
| Prior SOTA | 0.88 | 0.91 | 0.75 |
| zero-shot (Llama-2-70b) | 0.82 | 0.41 | 0.46 |
| zero-shot (Mixtral-8x7B) | 0.72 | 0.83 | 0.64 |
| zero-shot (GPT-3.5) | 0.74 | 0.80 | 0.59 |
| zero-shot (GPT-4) | 0.91 | 0.92 | 0.81 |
| *LLM-Match w/o classification head (Llama-3-8B-Instruct)* | 0.84 | 0.80 | 0.85 |
| *LLM-Match w/ classification head (Llama-3-8B-Instruct)* | 0.84 | 0.83 | **0.86** |

**Table 3.** Experimental Results on the SIGIR, TREC 2021, and TREC 2022 Datasets (Average).

| Model | AUROC |
| --- | --- |
| SciFive (encoder-decoder) | 0.5895 |
| BioBERT (dual-encoder) | 0.5952 |
| PubMedBERT (dual-encoder) | 0.5976 |
| SapBERT (dual-encoder) | 0.5933 |
| BioLinkBERT (cross-encoder) | 0.6176 |
| TrialGPT-Ranking (GPT-3.5) | 0.6582 |
| TrialGPT-Ranking (GPT-4) | 0.7979 |
| *LLM-Match w/o classification head (Llama-3-8B-Instruct)* | 0.7736 |
| *LLM-Match w/ classification head (Llama-3-8B-Instruct)* | **0.8010** |

*SIGIR, TREC 2021, and TREC 2022*: To evaluate LLM-Match, we also used patient summaries and clinical trials from three publicly available cohorts: the 2016 SIGIR test collection for patient-trial matching[24] and the 2021 & 2022 Clinical Trials (CT) tracks[25, 26] of the Text REtrieval Conference (TREC). The SIGIR cohort classifies patient-trial eligibility into three categories: irrelevant (would not refer this patient for this clinical trial), potential (would consider referring this patient to this clinical trial upon further investigation), and eligible (highly likely to refer this patient for this clinical trial). Similarly, the TREC cohorts define eligibility as irrelevant (the patient is not relevant for the trial in any way), excluded/ineligible (the patient has the condition that the trial is targeting, but the exclusion criteria make the patient ineligible), and eligible (the patient is eligible to enroll in the trial). To adapt the dataset for binary classification, we merged potential and eligible into a single class. Likewise, we combined excluded/ineligible and eligible into one class.

Table 1 summarizes key statistics for the four datasets, including the number of patients, clinical trials, and patient-trial pairs.

**Results**
We evaluate our model on four publicly available datasets: the n2c2 2018 cohort selection benchmark[23], the Special Interest Group on Information Retrieval (SIGIR) in 2016[24], and the 2021 and 2022 Clinical Trials (CT) tracks[25, 26] of the Text REtrieval Conference (TREC).

Table 2 compares LLM-Match with the basic baseline, prior state-of-the-art (SOTA) methods, and zero-shot[1] models. Among all approaches, LLM-Match with the classification head achieves the highest overall Macro-F1 score of 0.86, demonstrating its superior ability to balance performance across different classification categories. Macro-F1 is particularly crucial in clinical trial matching because of the inherent class imbalance, where the majority of patient records may not meet the eligibility criteria. A higher Macro-F1 score indicates that LLM-Match effectively reduces bias towards majority classes and ensures that minority cases, such as rare eligibility conditions, are also well-accounted for. Unlike prior SOTA methods that rely on proprietary models like GPT-4[13], our approach solely utilizes open-source large language models (LLaMA 3[27]) with even less model parameters (8B), proving that fine-tuned open models can achieve competitive, if not superior, results compared to expensive closed-source alternatives. This improvement can be attributed to the combination of retrieval-augmented generation (RAG) and fine-tuning with a structured classification head, which refines the model's decision boundaries and enhances its generalizability across various datasets. Furthermore, these results emphasize the feasibility of deploying transparent, cost-effective, and reproducible patient-matching solutions in real-world clinical applications, reducing reliance on commercial APIs that may impose usage restrictions or compliance risks.

**Table 4.** Experimental Results on the SIGIR, TREC 2021, and TREC 2022 Datasets (Separate).

| Model | Dataset | AUROC |
| --- | --- | --- |
| *LLM-Match w/o classification head (Llama-3-8B-Instruct)* | SIGIR | 0.7659 |
| *LLM-Match w/ classification head (Llama-3-8B-Instruct)* | SIGIR | 0.8100 |
| *LLM-Match w/o classification head (Llama-3-8B-Instruct)* | TREC 2021 | 0.7645 |
| *LLM-Match w/ classification head (Llama-3-8B-Instruct)* | TREC 2021 | 0.7829 |
| *LLM-Match w/o classification head (Llama-3-8B-Instruct)* | TREC 2022 | 0.7903 |
| *LLM-Match w/ classification head (Llama-3-8B-Instruct)* | TREC 2022 | 0.8102 |

Table 3 presents a comparison of the average AUROC across SIGIR[24], TREC 2021[25], and TREC 2022[26] datasets, further highlighting the robustness of LLM-Match. With an AUROC of 0.8010, LLM-Match with a classification head outperforms all other baselines, including TrialGPT-Ranking[2] with GPT-4[13], which previously set the highest standard in this domain. AUROC is a critical metric for evaluating the model's ability to distinguish between eligible and ineligible patients across varying decision thresholds. The superior AUROC score of LLM-Match suggests that the model is highly effective in capturing nuanced eligibility patterns, reducing false positives and false negatives in patient-trial matching. Notably, while GPT-4-based approaches have shown strong results, they are often limited by accessibility, cost, and regulatory concerns. Our work demonstrates that a well-optimized open-source model can match or even surpass the performance of closed-source alternatives, providing a scalable and transparent solution for clinical trial recruitment. The integration of retrieval-augmented generation ensures that LLM-Match efficiently selects the most relevant patient records, while the fine-tuned classification head enhances its decision-making capabilities. These results not only establish LLM-Match as a leading open-source alternative to proprietary models but also highlight the potential of democratizing AI-driven patient recruitment. Future improvements, such as reinforcement learning and advanced data preprocessing techniques, could further enhance the model's ability to generalize across diverse patient populations and clinical trial criteria.

Table 4 presents the results for SIGIR, TREC 2021, and TREC 2022 separately, showing performance variations across datasets. The LLM-Match model with a classification head consistently outperforms the version without it. These results indicate that the classification head significantly enhances model performance by enabling more effective feature extraction and better decision boundaries in the patient-matching task. The ablation study further confirms that adding a classification head enhances performance across datasets, including the n2c2 2018 Cohort Selection Challenge, by improving the model's ability to capture relevant features and distinguish complex patterns. The performance gains observed across multiple datasets highlight the robustness of the classification head in adapting to varied dataset structures and complexities.

**Discussion**
Our proposed method, LLM-Match, represents the first approach that integrates retrieval-augmented generation (RAG) with fine-tuning for patient matching in clinical trials. A key innovation of our approach is the addition of a classification head, which significantly enhances the model's predictive power by improving its ability to differentiate between eligible and ineligible patients. Unlike a generative head, which focuses on text generation and may introduce, the classification head is designed specifically for structured decision-making, ensuring precise, consistent, and task-specific outputs. This enhances predictability and interpretability, making it more suitable for applications requiring clear decision boundaries and accurate classification rather than free-form generation. Our results demonstrate that this enhancement leads to higher Macro-F1 scores and AUROC values compared to baseline models, indicating improved precision and recall in patient-trial matching. By refining decision boundaries, the classification head reduces ambiguity in model outputs, making it more robust and better suited for clinical applications, where reliability and accuracy are critical.

Further, the fine-tuning module helps the model adapt to specific tasks, such as patient matching in our case. Additionally, unlike many existing SOTA methods that rely on proprietary models such as GPT-4[13], LLM-Match exclusively utilizes open-source models like LLaMA 3[27] with even less model parameters (8B). This choice ensures greater accessibility, transparency, and adaptability, allowing researchers to fine-tune and customize models for specific clinical applications without reliance on proprietary systems. Open-source models also offer cost-efficiency and better data privacy control, addressing concerns related to patient data security in clinical settings. While closed-source models may benefit from extensive pretraining and proprietary optimizations, they pose limitations in reproducibility and interpretability. Our method demonstrates that smaller, open-source models can achieve competitive results, making them a viable alternative for resource-constrained environments while promoting a more collaborative and reproducible AI ecosystem in healthcare research.

However, while our approach improves overall performance, simply concatenating EHRs during prompt generation may introduce noise and reduce performance in certain cases. Merging diverse clinical notes without proper data preprocessing can lead to redundancy and inconsistencies, which may obscure critical eligibility information. Moreover, we do not account for time-dependent factors, and the token size is relatively small. Future iterations of LLM-Match should incorporate temporal dynamics and support larger, more realistic token sizes to better handle complex eligibility criteria.

Another limitation of this study is that it was not tested with real electronic health record (EHR) data, as the current datasets used are synthetic. Among them, n2c2 datasets are closer to real EHRs, whereas the other three datasets are more simplified representations. This limitation poses challenges for real-world applicability, as synthetic data may not fully capture the complexity, variability, and noise present in actual clinical records. Real EHRs often contain incomplete, ambiguous, or inconsistent information, which can significantly impact model performance. Additionally, structured and unstructured data elements in real-world EHRs may require more sophisticated data preprocessing and domain adaptation techniques to ensure effective patient-trial matching. Future work should focus on validating the model on real-world clinical datasets to assess its robustness, adaptability, and generalizability in practical healthcare settings.

Additionally, our current framework relies on a general fine-tuning process. While this approach is effective, reinforcement learning (RL) methods could further enhance the model's decision-making capabilities. By incorporating reinforcement learning, LLM-Match could dynamically adjust its weighting of eligibility criteria based on feedback, potentially leading to more nuanced and adaptive matching decisions. By formulating patient-trial matching as a sequential decision-making process, the model can learn from feedback, refining its ability to identify eligible patients while minimizing false positives and false negatives. Additionally, RL can enhance adaptability to evolving clinical trial criteria by continuously updating its policy based on new data, ensuring improved generalization across different trials and healthcare systems. Furthermore, integrating RL could enable more efficient handling of ambiguous or borderline cases, where traditional rule-based or supervised learning methods may struggle.

## Conclusions

In this study, we introduced LLM-Match, a novel framework that combines retrieval-augmented generation and fine-tuning to improve patient-trial matching. Our evaluation on multiple benchmark datasets, including n2c2[23], SIGIR[24], TREC 2021[25], and TREC 2022[26], demonstrates that LLM-Match achieves superior performance over existing methods. The integration of a classification head further enhances the model's predictive accuracy, underscoring the importance of tailored deep-learning architectures for patient matching. Unlike previous SOTA models that rely on proprietary LLMs such as GPT-4[13], our approach exclusively leverages open-source models, proving that they can achieve comparable and even superior results when properly fine-tuned. This advancement offers a scalable, transparent alternative to black-box AI, enhancing accessibility and reproducibility for clinical research.

## Acknowledgments

This work is supported by a grant from the National Institute of Health (NIH) NIGMS (R00GM135488).

## Code Availability

The code for this work LLM-Match is publicly available at https://github.com/bioIKEA/LLMMatch.